%% file: submission/main_arxiv.tex
\documentclass[sigconf,nonacm]{acmart}
\usepackage{xspace}
\usepackage{dsfont}
\usepackage{setspace}
\usepackage{enumitem}
\usepackage{setspace}

\AtBeginDocument{%
  }

\setcopyright{acmlicensed}
\copyrightyear{2018}
\acmYear{2018}
\acmDOI{3705328.3748050}
\acmConference[Conference acronym 'XX]{XXX}{June 03--05,
  2018}{Woodstock, NY}
\acmISBN{978-1-4503-XXXX-X/2018/06}
\newcommand{\pinfm}{PinFM\xspace}
\newcommand{\inv}{^{\raisebox{.2ex}{$\scriptscriptstyle-1$}}}

\begin{document}

\title{\pinfm: Foundation Model for User Activity Sequences at a Billion-scale Visual Discovery Platform}

\author{Xiangyi Chen}
\email{xiangyichen@pinterest.com}
\affiliation{%
  \institution{Pinterest}
  \city{San Francisco}
  \state{CA}
  \country{USA}
}

\author{Kousik Rajesh}
\email{krajesh@pinterest.com}
\affiliation{%
  \institution{Pinterest}
  \city{San Francisco}
  \state{CA}
  \country{USA}
}

\author{Matthew Lawhon}
\email{mlawhon@pinterest.com}
\affiliation{%
  \institution{Pinterest}
  \city{San Francisco}
  \state{CA}
  \country{USA}
}

\author{Zelun Wang}
\email{zelunwang@pinterest.com}
\affiliation{%
  \institution{Pinterest}
  \city{San Francisco}
  \state{CA}
  \country{USA}
}

\author{Hanyu Li}
\email{hanyuharryli@pinterest.com}
\affiliation{%
  \institution{Pinterest}
  \city{San Francisco}
  \state{CA}
  \country{USA}
}
\author{Haomiao Li}
\email{haomiaoli@pinterest.com}
\authornotemark[1]
\affiliation{%
  \institution{Pinterest}
  \city{San Francisco}
  \state{CA}
  \country{USA}
}

\author{Saurabh Vishwas Joshi}
\email{sjoshi@pinterest.com}
\affiliation{%
  \institution{Pinterest}
  \city{San Francisco}
  \state{CA}
  \country{USA}
}

\author{Pong Eksombatchai}
\email{pong@pinterest.com}
\affiliation{%
  \institution{Pinterest}
  \city{San Francisco}
  \state{CA}
  \country{USA}
}

\author{Jaewon Yang}
\email{jaewonyang@pinterest.com}
\affiliation{%
  \institution{Pinterest}
  \city{San Francisco}
  \state{CA}
  \country{USA}
}
\author{Yi-Ping Hsu}
\email{yhsu@pinterest.com}
\authornote{Work done at Pinterest.}
\affiliation{%
  \institution{Pinterest}
  \city{San Francisco}
  \state{CA}
  \country{USA}
}

\author{Jiajing Xu}
\email{jiajing@pinterest.com}
\affiliation{%
  \institution{Pinterest}
  \city{San Francisco}
  \state{CA}
  \country{USA}
}

\author{Charles Rosenberg}
\email{crosenberg@pinterest.com}
\affiliation{%
  \institution{Pinterest}
  \city{San Francisco}
  \state{CA}
  \country{USA}
}


\begin{abstract} 
User activity sequences have emerged as one of the most important signals in recommender systems. We present a foundational model, \pinfm, for understanding user activity sequences across multiple applications at a billion-scale visual discovery platform. We pretrain a transformer model with 20B+ parameters using extensive user activity data, then fine-tune it for specific applications, efficiently coupling it with existing models. While this pretraining-and-fine-tuning approach has been popular in other domains, such as Vision and NLP, its application in industrial recommender systems presents numerous challenges. The foundational model must be scalable enough to score millions of items every second while meeting tight cost and latency constraints imposed by these systems. Additionally, it should capture the interactions between user activities and other features and handle new items that were not present during the pretraining stage.

We developed innovative techniques to address these challenges. Our infrastructure and algorithmic optimizations, such as the Deduplicated Cross-Attention Transformer (DCAT), improved our throughput by 600\% on Pinterest internal data. We demonstrate that \pinfm can learn interactions between user sequences and candidate items by altering input sequences, leading to a 20\% increase in engagement with new items. \pinfm is now deployed to help improve the experience of more than half a billion users across various applications.

\end{abstract}

\ccsdesc[500]{Information systems~Recommender systems}

\keywords{Sequential Recommendation, Recommendation System}


\maketitle

\section{Introduction}\label{sec:intro}
\input{submission/intro}

\section{Related Work}\label{sec:related}
\input{submission/related}

\section{Methodology}\label{sec:method}
\input{submission/method}

\section{Efficiency}\label{sec:infra}
\input{submission/infra}

\section{Experiments}\label{sec:experiments}
\input{submission/experiments}

\section{Conclusion}\label{sec:conclusion}
We presented \pinfm, a billion-scale visual discovery platform's first foundational model for user activity sequences. We pretrain a large transformer with over 20 billion parameters using years of user activity data. Subsequently, we fine-tune \pinfm with downstream ranking models for each application. During the fine-tuning stage, \pinfm learns the interactions between user activity sequences and candidate items while retaining the knowledge acquired during pretraining.  We enhanced \pinfm's training and serving efficiency by introducing the Deduplicated Cross-Attention Transformer and quantizing the large embedding tables. As a result, we successfully deployed \pinfm across multiple applications without incurring significant cost increases. For future work, we aim to further improve the training and serving efficiency of \pinfm and incorporate more diverse entities into the input sequences. As the platform's first foundational model, \pinfm establishes a new paradigm for the company's recommender systems.


\bibliographystyle{ACM-Reference-Format}
\bibliography{submission/base}

\end{document}

%% file: submission/intro.tex

Recommender systems (RecSys) provide relevant and engaging content to users in many social media platforms.
For personalization, recommender systems need to comprehend users' past interaction histories, often referred to as user activity sequences. Early methods focused on extracting the attributes of the items users interacted with~\cite{FM10}. More recent approaches have recognized the importance of the sequential nature of user activities and are therefore focused on developing model architectures specialized for sequences, such as transformers. Recent studies have also found that utilizing a large model to understand user activity histories enhances predictive performance and user engagement metrics~\cite{Scalinglaw24, HSTU24}.

In this paper, we aim to build a foundational model for understanding user activity sequences for multiple RecSys applications at a billion-scale visual discovery platform. We pretrain a large foundational model with a massive amount of user activity sequence data collected across multiple applications. For each downstream application, we fine-tune the foundational model with application-specific training data and application-specific features. This approach allows for training a very large model efficiently because the pretraining can be done for multiple applications. It also allows for integrating the foundation model easily with existing application model. For these reasons, this pretraining-and-fine-tuning approach has become highly popular in both large language models and computer vision tasks.


There are many challenges in building large foundational model for understanding user sequences for multiple recommendation applications at industrial scale. First, since the foundational model needs to be fine-tuned with a downstream recommender system model, it is crucial to keep the training and serving cost and latency at a manageable level. Recommender system models need to process the request within hundreds of milliseconds and score millions of items every second ~\cite{Transact23}. Developing a large model to satisfy this latency budget is very challenging. Second, applications would have different sequences and application-specific features about the candidate item and the user. It is crucial to capture the interaction between the user activities and other features used in the application~\cite{DIN18}. Third, foundational models should generalize to new items not present in the pretraining data, as users keep introducing new items everyday.

Though the effectiveness of large sequence models is well known in industrial recommender systems~\cite{Tiger23, HSTU24}, to our knowledge, there is no existing literature regarding developing foundational models for user sequences to be integrated with multiple downstream recommender system models. Instead, large sequence models such as HSTU~\cite{HSTU24}, TIGER~\cite{Tiger23}, and TWIN-V2~\cite{TWINV224} are often used as standalone models trained for specific applications, which can become prohibitively expensive if applied across many applications. An alternative approach to developing large foundational models is to use them as teacher models in knowledge distillation~\cite{KDFMMeta25, KDGoogleGap24}. Unfortunately, this method significantly slows down experimentation, as it requires waiting for the production model to learn from the new teacher model, a process that can take months~\cite{KDGoogleGap24}.

We present \pinfm, a foundational model for user activity understanding. We pretrain a large scale transformer using a massive dataset of user activity sequences. In particular, we pretrain a 20 billion parameter model with 2 years of user activity history. To expedite pretraining, we avoid using high-dimensional features for each user activity, such as pretrained item embeddings. Instead, we rely on categorical features like item IDs and action types, learning embeddings for each ID.
Once pretraining is complete, we integrate the pretrained transformer and embedding tables within the recommender system models for multiple applications. Each downstream ranking model uses this pretrained transformer and embedding table to encode user activity sequences and combine it with other existing features. To address the first challenge of training / serving cost, we developed some novel optimizations technique such as the Deduplicated Cross-Attention Transformer (DCAT) and int4 quantized embeddings. These improvements allow us to deploy \pinfm to the two most important recommender systems, the Home Feed and Related Items Feed, with neutral cost and marginal latency increases. To address the second challenge of capturing interactions between the user sequence and other features, we explored different ways of creating the input sequence for the foundational transformer, so that the transformer can capture interactions via cross attention. To address the last challenge of handling unseen items, we added modeling techniques such as leveraging item features and additional dropout for fresh items. 

The summary of our contributions is as follows:
\begin{itemize}
    \item \textbf{Foundational paradigm for user activity understanding}: We propose a pretraining-finetuning approach for user activity sequence models. Our pretraining method can handle years of user activity sequences efficiently by avoiding high dimensional features. The pretrained model can be easily integrated with multiple downstream models, as user activity sequences are a common signal for most recommender systems. To the best of our knowledge, this is the first foundational model for understanding user sequences in the industry.
    \item \textbf{Efficiency improvements for cost-neutral integration with downstream models}: We developed several efficient computational techniques for \pinfm. For instance, \pinfm processes the user activity sequence just once, reusing the results for all candidates during the ranking stage. Additionally, we implement aggressive quantization for the large embedding tables. These techniques allow \pinfm to be seamlessly integrated into a wide range of downstream recommender models. This integration does not affect latency or serving costs, allowing downstream models to continue utilizing other features as they normally would.
    \item \textbf{Input sequence engineering for learning user-candidate interactions:} We conducted experiments to demonstrate that \pinfm can capture the interactions between the user activity sequences and other features, in particular candidate item features. This flexibility enables us to design input sequences tailored to different applications to meet specific needs. For example, in the Home feed application, where there are many fresh items, we incorporated item features into the input sequence to address the cold start problem. This adaptability is crucial for making a significant impact across multiple applications.
    \item  \textbf{Sharing lessons from live traffic experiments:} We deployed \pinfm to the platform's two most crucial  recommender systems, the Home feed and the Related Items feed. Through extensive ablation studies conducted in both offline and online tests, we identified the most impactful techniques. 
\end{itemize}

The rest of the paper is organized as follows. Sec.~\ref{sec:related} discusses related work. Sec.~\ref{sec:method} describes \pinfm, and  Sec.~\ref{sec:infra} presents infrastructure optimization. Sec.~\ref{sec:experiments} presents offline and online experimental results. We conclude in Sec.~\ref{sec:conclusion}.

%% file: submission/related.tex
We discuss related work in two areas: User sequence understanding models and Foundation model for recommender systems.

\textbf{User sequence models:} Users' past interaction histories have been valuable signals for Recommender systems from early days~\cite{FM10}. Transformers, by capturing the interactions in sequential data via self-attention, revolutionized user sequence understanding models~\cite{KangM18, Pinnerformer22, Transact23, TWINV224}. One approach is to use sequence models to predict the next item that the user would interact with. These methods have shown promising results in academic benchmarks~\cite{GRU4REC16, MultiBehaviorSASRec24, UniS4Rec22} and in industry~\cite{Tiger23, HSTU24}. However, they can only handle sequences; it is not straightforward for them to leverage features that are designed for non-sequential models like DCN or DLRM~\cite{DLRM19, DCNV221}.

Another line of methods is developing a sequence encoder within an existing recommender system models. Methods like TransAct~\cite{Transact23}, TWIN~\cite{TWIN23, TWINV224} developed a transformer to encode the user activity sequences. These transformers are trained with the downstream application models with other application-specific features. We aim to combine both approaches. We pretrain a sequence-only model to train user activities across applications, but we fine-tune \pinfm with existing features in the downstream models.

\textbf{Foundation models for recommender systems} Foundation models for recommender systems are large models trained across various applications to handle multiple recommendation tasks. For a detailed survey, please refer to \cite{FMRecSysSurvery24}. One major trend is to use large language models as a backbone for a foundation model. Unis4Rec~\cite{UniS4Rec22}, RecFormer~\cite{Recformer23} and PTUM~\cite{PTUM20} showed promises in academic benchmarks, and M6Rec~\cite{M6Rec22} and 360Brew~\cite{360Brew25} did in industrial applications. Despite the impressive offline results, these models are computationally heavy due to processing long prompts that describe the full feature vector in text. Therefore, we decided to develop a foundation model that encodes user sequences using a sequence of item IDs, which is significantly more compact than text.


%% file: submission/method.tex
As mentioned earlier, our framework contains two stages, pretraining a user sequence model and fine-tuning with downstream ranking models, an illustration of the two-stage paradigm can be found in Figure \ref{fig:frm_diagram}. In this section we explain our method in more details. 
{\captionsetup{belowskip=-10pt}
\begin{figure*}
    \centering
    \includegraphics[width=\linewidth]{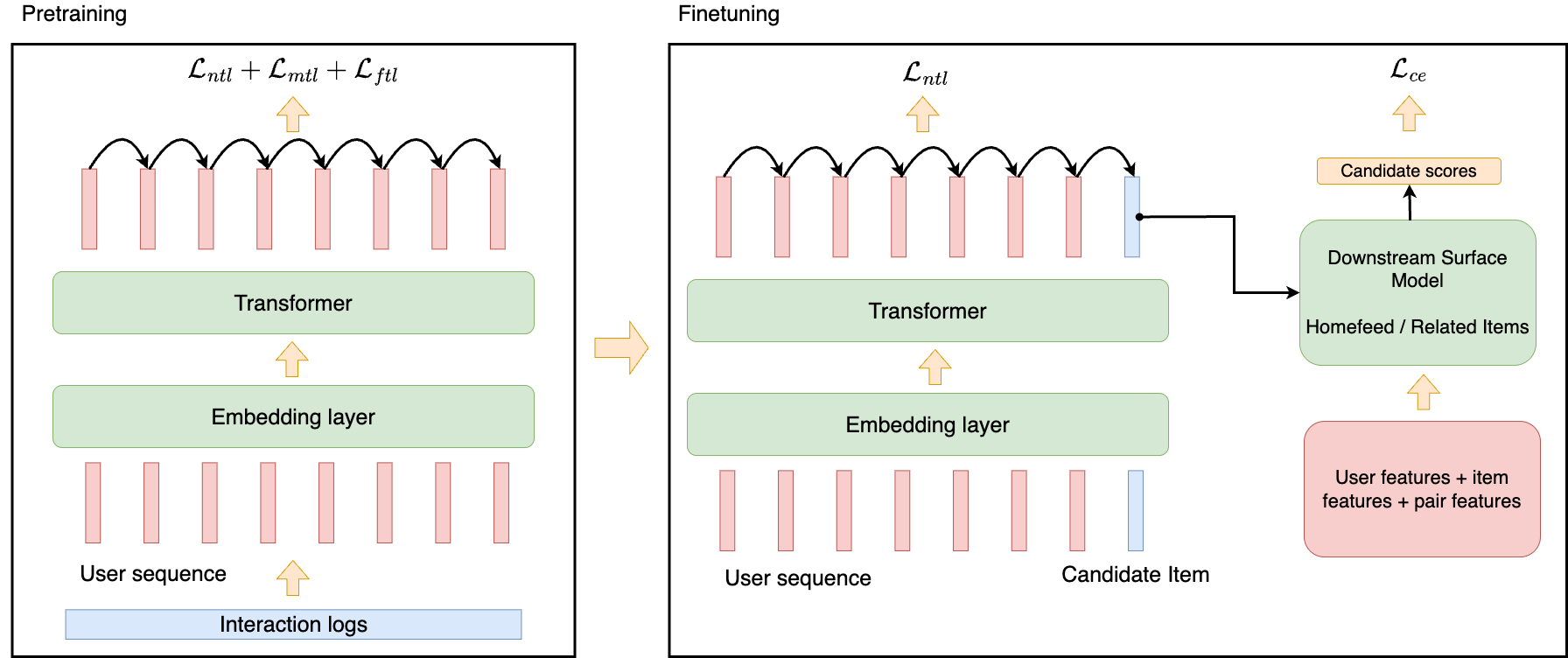}
    \caption{Training and fine-tuning of \pinfm}
    \label{fig:frm_diagram}
\end{figure*}
}
\subsection{Pretraining}
In the pretraining stage, we collect user sequence data and use a few variants of next token prediction loss commonly used in language models. In our framework, the backbone model architecture can be any transformer-based decoder-only models, and we choose the GPT2 architecture \cite{radford2019language} with Pre-LN \cite{xiong2020layer} in all our experiments to balance performance and latency requirements.  We also tried HSTU \cite{HSTU24} architecture and got similar results with GPT2.  In the next, we discuss how we construct data and modified loss functions used in large language models to address the challenges unique to recommender systems.

\textbf{Data Construction.} We collect each user's historical actions on the platform and sort them in ascending order by timestamp, capping the maximum sequence length at 16{,}000. We denote a user’s full action sequence as $\mathbf{S} = [S_1, S_2, \ldots, S_N]$, where each element $S_i$ represents the $i$-th event. Specifically, $S_i = [t_i, a_i, v_i, id_i]$ is a tuple consisting of the timestamp $t_i$, the action type $a_i$, surface type $v_i$, and the identifier $id_i$ of the item on which the action was performed. During training, we cut the user sequences into non-overlapping segments of length $L$ with $L$ being a few hundred.

\textbf{Modeling and Training Objectives.} In training, given a possibly truncated user sequence $\mathbf{S}$ with length $m$, we map each item identifier $id_i$ to an embedding vector $\mathbf{E}_i = \operatorname{emb}(id_i)$, forming the embedding matrix $\mathbf{E} = [\mathbf{E}_1, \mathbf{E}_2, \ldots, \mathbf{E}_m]$. The surface and action type sequences are also converted to embedding matrices $\mathbf{V} = [\mathbf{V}_1, \mathbf{V}_2, \ldots, \mathbf{V}_m]$ and $\mathbf{A} = [\mathbf{A}_1, \mathbf{A}_2, \ldots, \mathbf{A}_m]$ by an embedding table.  Next, we apply a pointwise MLP with an $l_2$ normalization layer, denoted $\phi_{\mathrm{in}}$, to produce an input to the backbone model $M$. The final hidden states from $M$ are then passed through another pointwise MLP with normalization, $\phi_{\mathrm{out}}$, yielding the user representation sequence $\mathbf{H}$. Formally speaking, 
\begin{equation} \label{eq: forward}
\mathbf{H} = [\mathbf{H}_1, \mathbf{H}_2, \ldots, \mathbf{H}_m] = \phi_{\mathrm{out}}(M(\phi_{\mathrm{in}}(\mathbf{E + V + A})))
\end{equation}

For our training objectives, we build upon language model pretraining~\cite{GPT3_20} and introduce novel objectives specifically designed for recommender systems. We begin with a commonly used loss function for user sequence modeling—the next token prediction loss, which is predicting the next item based on the user's activities leading up to that point. However, previous studies~\cite{Pinnerformer22, sessionrec25} have shown that next token prediction alone is insufficient to capture the evolution of users' interests over time. Since users can have multiple interests and switch between short-term and long-term interests, it is beneficial to aim at predicting multiple tokens simultaneously (see Sec.~\ref{sec:ablations} for more details). To address this, we incorporate multiple types of future token predictions during pretraining.

There are additional differences between pretraining user activity models and language models. In user activities, we have action types, with varying levels of importance among them. Additionally, the vocabulary size for user activities—the number of unique items—can reach billions, making it computationally expensive to calculate the prediction probability for every token in the vocabulary. To address these challenges, we formulate our pretraining objectives as follows.


1. We adopt infoNCE-based loss \cite{oord2018representation} due to large vocabulary size. For each user representation $\mathbf{H}_i$, when there is a positively engaged item after $i$th position in the sequence, whose embedding is obtained by $z_{i+1} = \psi (\operatorname{emb} (id_{i+1}))$ with $\psi$ being another MLP followed by an $l_2$ normalization layer,  the loss function below can be applied
\begin{equation} \label{eq:loss}
    l(\mathbf{H}_i, z_{i+1}) = - \log \frac{\exp(sim(\mathbf{H}_i, z_{i+1})/\tau)}{\exp(sim(\mathbf{H}_i, z_{i+1})/\tau) + \sum_{k=1}^{K} \exp(sim(\mathbf{H}_i, z^-_{k})/\tau)}
\end{equation}
where we choose $sim(a, b) = a^Tb$ being the inner product,  $\tau$ being the temperature parameter and we make it learnable with a small initial value. 
$z_k^-$ are $K$ embeddings sampled in-batch excluding items positively engaged by the same user, these are treated as negative samples in the contrastive loss. 

2. As indicated in equation $\ref{eq:loss}$, the loss is only applied when there is a future positively engaged item for $\mathbf{H_i}$. This can drastically reduce computation compared with predicting tokens at every position. Let $A_{pos}$ denote the set of predefined positive actions and $\mathds{1}[]$ be the indicator function, the Next Token Loss can be defined below
$$
\mathcal{L}_{ntl} = \sum_{i=1}^{m-1} l(\mathbf{H}_i, z_{i+1}) \mathds{1}[a_{i+1} \in A_{pos}]
$$

3. In addition to predicting each positively engaged next token, we also want the model to predict future tokens in a window of length $L'$ since user interests tend to be consistent in a short period of time. Thus, we add an extra Multi Token Loss defined as below. 
$$
\mathcal{L}_{mtl} = \sum_{i=1}^{m-1} \sum_{j=i+1}^{i+L'} l(\mathbf{H}_i, z_{j})  \mathds{1}[a_{j} \in A_{pos}]
$$
where $z_{j}$ is a positive item within the window. In practice, we also subsample the loss to reduce computation cost.

4.  Our model will eventually be used in downstream ranking models, and these ranking models often use a real-time sequence with max length $L_d$ smaller than $L$ due to latency constraints. The losses $\mathcal{L}_{ntl}$ and $\mathcal{L}_{mtl}$ put approximately equal weights on predicting future tokens for each input sequence length, but we want the model to be more predictive especially when the sequence length is close to the downstream input sequence length $L_d$. To achieve this, we  add another loss to predict a future window of positive tokens on $\mathbf{H}_{L_d}$, we term this the Future-Token Loss ($\mathcal{L}_{ftl}$).  
$$
\mathcal{L}_{ftl} = \sum_{j=L_d+1}^{L_d+L'} l(\mathbf{H}_{L_d}, z_{j}) \mathds{1}[a_{j} \in A_{pos}]
$$
This is similar to instruction fine-tuning in large language models, where we present a task to the model by providing a complete input sequence (instruction) and prompting it to generate the output.

\subsection{Fine-tuning}
Once the model is pretrained on the user sequence data, we fine-tune it with downstream ranking models. In the ranking stage, capturing the interaction between user features (including \pinfm) and candidate item features is crucial. We will discuss how to achieve this without significantly increasing computational costs.

\textbf{Ranking model integration.}
 Most of our ranking models are classification models with feature crossing layers like DLRM~\cite{DLRM19} or DCN~\cite{DCNV221}. In a request for a specific user, we rank the candidate items returned by retrieval models. For each candidate item, the ranking model applies modules that process user features, candidate item features, and interaction features. The outputs of these modules are then aggregated to predict the probabilities of various user actions for the candidate item. An example of ranking models can be found in \cite{xia2023transact} with more details. 
 
 When integrating with ranking models, we add the pretrained model as a module for the user activity sequence. One key difference compared to pretraining comes from the presence of the candidate item. There are two ways to fuse the user activity sequence and the candidate item.
 

1. Late fusion. The module can take a user sequence as input to produce a user embedding, which will be aggregated with outputs from other modules later. This setting makes the input sequence to \pinfm identical to the pretraining task because we do not consider the candidate item. We can easily cache the output of \pinfm for candidate items related to the same request because the user activity sequence remains identical. However, the downside is that \pinfm won't be able to contextualize the user activity sequence for each candidate item.

2. Early fusion. The candidate item is appended to the user sequence used as input to the user sequence module. In this case, we can output a user embedding and another embedding crossing the user information with candidate information~\cite{HSTU24}. Due to its stronger prediction ability (See Sec.~\ref{sec:ablations}), we choose this way. With this approach, the input sequence to \pinfm varies for each candidate item within the same request, but we will show in Sec~\ref{dcat} that we can still efficiently cache intermediate results for the candidates in the same request.

It is noteworthy to mention that for the candidates to be ranked, there are typically some extra embedding features available, such as content embedding, graph-based embedding. For an extra candidate embedding, we also optionally project and sum it with the projected candidate id embedding as the user sequence module input, an extra loss to align the projected embedding into the user sequence module input space is added in this case.

To let the pretrained model adapt to the sequence distribution of downstream ranking models faster, we can optionally add the losses $\mathcal{L}_{ntl}$ and $\mathcal{L}_{mtl}$ in the fine-tuning stage in addition to the downstream ranking model losses. Also, we set the learning rate of the pretrained module to be around 1/10 of the ranking model learning rate to keep the pretrained information from being overwritten.

To further align the learning target of the user sequence module with the downstream model, we also apply the ranking losses of the downstream model to the user sequence module outputs, together with an MSE loss to align the predictions given by our user sequence module and the final ranking model predictions. We found that this helps boost the performance of downstream ranking models.

\textbf{Handling Cold-start items.}
As it is commonly known, id embeddings only perform well for items with enough interaction history. When we add a module that is strongly dependent on the candidate item id embedding, the model performance on cold-start items could gets hurt. To recover the performance, we found the techniques below to be helpful

1. Candidate item id randomization. During fine-tuning, we randomly choose an id for the candidate item 10\% of times. This can simulate cold-start situations since the cold-start item ids are randomly generated and not seen by the model during training.  

2. Dropout on module outputs for fresh candidate items. During fine-tuning, we add dropout to the outputs of our pretrained module if the candidate item is fresh, i.e., generated within T days of the training sample request time. With such change, the model can hopefully rely less on outputs of the module that are derived from id embeddings for fresh candidates.

More details on these techniques along with others that helped cold-start performance can be found in Sec.~\ref{sec:ablations}.

\subsection{Discussion on design choices}

\textbf{Unidirectional vs Bidirectional modeling.} 
During pretraining we prefer unidirectional transformers for similar reasons as large language models. Compared to bidirectional transformers, unidirectional models trained using next token prediction objectives hold more predictive power, allowing the model to generalize to new user sequences. In fine-tuning, if we eliminate causal modeling in favor of bidirectional attention, we note a 1.4\% drop in evaluation metrics. We hypothesize that this drop is due to the distribution shift between pretraining and fine-tuning. 
Further, the use of causal attention allows us to significantly optimize the attention mechanism for data patterns we observe, as explained in section  \ref{dcat}.
\\
\textbf{ID vs pretrained embeddings in sequence.} 
Instead of using ID-based sequence representation, a straightforward alternative is to use pretrained content-based or engagement-based embeddings. However, using pretrained embeddings will put more pressure and challenges on infrastructure since the data volumn for both storage and serving needs to be hundreds of times larger for the user sequence features. Also, it was shown in \cite{hsu2024taming} that pretrained ID embedding can transfer well to downstream models.  Therefore, we choose ID-based representation for its flexibility and scalability. 


%% file: submission/infra.tex
The framework's core components are transformers and large embedding tables. We allocated significant engineering resources to optimize both components, as their efficiency is critical for the project's feasibility from an infrastructure cost standpoint.

\subsection{Deduplicated Cross-Attention Transformer (DCAT)}
\label{dcat}
To deploy our large transformer architecture in online ranking systems handling a few million queries per second, we capitalize on a key data pattern. There are significantly fewer unique user sequences compared to the number of candidates that are scored -- the ratio is 1:1000 during serving and 1:10 during training. The transformer computation in our model architecture can be separated into two components: (1) the context component, which involves applying the transformer to a user's historical action sequence, and (2) the crossing component, where each item to be scored is evaluated in conjunction with the user history. The context component is computed only once per user, and the keys and values of each transformer layer are stored as KV cache. We then use the context KV cache to perform cross-attention with the entire list of candidate items.  This optimization is implemented in both the training and serving phases with custom kernels using OpenAI Triton, and helped us achieve a 600\% increase in serving throughput and 200\% increase in training throughput over regular self attention implemented with FlashAttention \cite{fav2}.

Now we explain DCAT in more detail. We start with the context component, which produces transformer KV cache for the crossing component using a de-duplicated user sequence. Let $\mathbf{X}^{(0)} \in \mathbb{R}^{B \times L_d \times d}$ be a batch of user sequence embeddings as input of the transformer, and $\Psi$ be an invertible deduplication operation in the batch dimension. We first deduplicate the sequence input by 
$$\mathbf{X}_{u}^{(0)} = \Psi(\mathbf{X}^{(0)}) \in \mathbb{R}^{B_u \times L_d \times d}$$
and $\mathbf{X}_{u}^{(0)}$ is the deduplicated batch, typically $B_U = B/16$. 
Denote $\mathbf{X}_{u}^{(l-1)}$ as the input to $l$-th layer of the transformer, then we can compute $\mathbf{X}_{u}^{(l)}$ by 
\begin{equation}
\begin{aligned}
\mathbf{Q}_u^{(l)} &= W_qX_u^{(l-1)}, \quad \mathbf{K}_u^{(l)} = W_kX_u^{(l-1)}, \quad \mathbf{V}_u^{(l)} = W_vX_u^{(l-1)} \\
    &\mathbf{X}_{u}^{(l)} = g(\text{Attention}(\mathbf{Q}_u^{(l)},  \mathbf{K}_u^{(l)},  \mathbf{V}_u^{(l)}), X_u^{(l-1)})
\end{aligned}
\end{equation}
with $g()$ being some processing function that is irrelevant to the context here and $\text{Attention}()$ being the standard softmax attention.
We save $\mathbf{K}_u^{(l)}$ and $\mathbf{V}_u^{(l)}$ as KV cache to be used later. 
In the crossing component, we have a batch of candidate embeddings $\mathbf{X}_{c}^{(0)} \in \mathbb{R}^{B \times 1 \times d}$ as transformer input. For each layer, we apply the inverse of the deduplication operation $\Psi^{-1}$ to the KV cache and concatenate them with the candidate KV, before performing cross attention. Given $\mathbf{X}_{c}^{(l-1)}$ as $l$-th layer input, $\mathbf{X}_{c}^{(l)}$ can be written as
\begin{equation}
    \begin{aligned}
     \mathbf{Q}_c^{(l)} &= W_qX_c^{(l-1)}, \quad \mathbf{K}_c^{(l)} = W_kX_c^{(l-1)}, \quad \mathbf{V}_c^{(l)} = W_vX_c^{(l-1)}  \\
    \mathbf{X}_{c}^{(l)} &= g(\text{Attention}(\mathbf{Q}_c^{(l)}, \Psi\inv(\mathbf{K}_u^{(l)})\: ||\: \mathbf{K}_c^{(l)},  \Psi\inv(\mathbf{V}_u^{(l)})\: ||\: \mathbf{V}_c^{(l)}), X_c^{(l-1)})
    \end{aligned}
\end{equation}
and \textbf{||} operator denotes concatenation along the sequence dimension. 

In training, $\Psi$ is implemented using Pytorch and in serving, the inference server uses pointers to perform the same deduplication. In both training and serving, 
$\Psi\inv$ is 
implemented using a Triton kernel and serves as a "broadcasting" operation that uses the unique sequence indices from $\mathbf{X}^{(0)}$ to invert the deduplication operation.

GPUs typically perform much better when input tensors are aligned to powers of 2. We further optimize DCAT by maintaining the sequence length to be 256 at all times, eliminating all concatenations, and instead replacing the oldest tokens in KV cache with the candidate KV and rotating the attention mask. During serving, we also skip the self attention operation associated with the calculation of $X_u^{(l)}$ for the last layer, since this is only used for loss computation. Together, these increase our throughput by 25\% additionally over the 600\% improvement from DCAT.

\subsection{Large Embedding Table}
We adopted a straightforward setup for both the training and serving of our large embedding tables. On the training side, we utilized TorchRec \cite{10.1145/3523227.3547387} to distribute our tables across multiple GPUs. On the serving side, the tables are hosted on a single CPU host, separate from the GPU inference host, as its size is manageable within the memory limits of a single CPU. While the training setup is scalable, the serving setup is less flexible due to the memory constraints of each CPU host and the significant communication costs incurred when the sequence of embeddings are transferred to the GPU inference host. To address this, we investigated various compression techniques to decrease data size, ultimately adopting a post-training quantization strategy.

\textbf{Embedding quantization:} Within the proposed foundation model, the vast majority of the parameter increase is attributed to the large embedding table.  For example, in our standard setup, for each item id $id_i$, we obtain a 256 dimensional float16 embedding vector through looking up 8 sub embedding tables, each with 80 million rows and 32 dim, which helps mitigate hash collision: 
$\mathbf{E}_i = \operatorname{emb}(id_i) = \bigotimes_{j = 0}^{7}{\operatorname{emb_j}(\operatorname{hash_j}(id_i))}$ in which we use $\bigotimes$ to denote concatenation of the the 8 sub embeddings. 
The embedding tables overall adds 20 billion trainable parameters. While vastly increasing the model capacity, it also poses a significant challenge for serving. 

We therefore performed quantization after training the model. Specifically, we used FBGEMM\cite{fbgemm}'s n-bit min-max quantization kernels as backbone and performed post-training quantization (PTQ) across all shards of the embedding table, effectively converting each 32-dim fp16 vector to 32 int8/int4 + 1 fp16 scale value + 1 fp16 bias value, and then performed bitpacking to fp32 to reuse torch native embedding API. In an int4 setting, this compresses each vector from 512 bit to 160 bit, and reduces the embedding table to 31.25\% of its original size, and cuts CPU host cost proportionally.

At serving time, we serve the compressed table on CPU hosts as in Figure \ref{fig:serving_infra} while on GPU, we optimize decoding by using a custom Triton kernel that combines bit unpacking and FBGEMM dequantization. In online serving tests, we observed negligible GPU forward pass latency overhead, and overall inference API call latency was reduced by 7\% due to lowered embedding IO load. 

We quantify the effect of quantization by calculating the L2 norm of the value deviation between original embedding and quantized embedding vs the L2 norm of original embedding, we observed 0.45\% at int8 quantization, and 7.8\% at int4.

In offline evaluation of Save metrics, we observed that int8 quantization had a negligible effect while int4 brings 0.06\% drop versus full precision model, which is still quite minor. We proceeded with an online A/B test with int4 PTQ against full precision model. Over two weeks, we observed statistically neutral performance and validated int4 quantization as a viable optimization.
{\captionsetup{belowskip=-15pt}
\begin{figure}
    \centering
    \includegraphics[width=\linewidth]{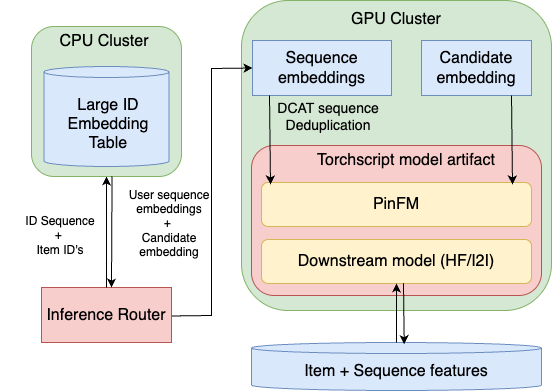}
    \caption{Serving Infrastructure for \pinfm}
    \label{fig:serving_infra}
\end{figure}
}
\subsection{Serving Infrastructure}
Figure \ref{fig:serving_infra} shows the serving infrastructure design for \pinfm. Notably, we separate our large ID embedding table from the dense model and serve it on a CPU cluster. This allows us to scale our embedding table and model size without having to share the limited GPU memory available during inference.
The inference router is responsible for fetching the relevant ID embeddings for the user sequence and candidate item from the CPU cluster and passing it to the GPU cluster for inference. The sequence embedding is de-duplicated here before the model forward using DCAT. The final token output from \pinfm is then passed to the downstream model for score prediction.

%% file: submission/experiments.tex
In this section, we compare \pinfm to baselines, evaluate different designs for \pinfm, and explore the optimal input sequence engineering for downstream applications. We also show that \pinfm delivers strong metric wins in online A/B tests in multiple applications.

\subsection{Offline Experiments}\label{sec:ablations}
For offline experiments, we modify \pinfm and pretrain it for different ablations. We then fine-tune with  downstream applications and report the downstream application metrics.

\textbf{Dataset and metrics:} As described in Sec~\ref{sec:method}, we use a pretraining dataset of the last 2 years of user activities. For fine-tuning applications, we use the platform's Home Feed (HF)~\cite{xia2023transact} and Related Items (Item2Item, I2I). When a user comes in to the platform, the Home Feed recommends the most relevant and engaging items. When the user clicks on a particular item, I2I shows other items that are related to the clicked item. Both recommender systems consist of retrieval models and ranking models, and we apply \pinfm to the ranking models. For fine-tuning data, we use 3 weeks of HF / I2I data.

HF and I2I are the two most important ranking models for user engagement at the platform. Both models are point-wise multi-task learning models to predict probabilities of various user actions, like Save (save to users' collections), Click, Share, Hide, etc. Both models use item features, user features, and context features as input, and both surfaces aim to recommend personalized feed and to serve fresh content. The difference between HF and I2I is that HF recommendations are diverse and exploratory, whereas I2I has an emphasis on relevance between candidate items and the query item and focuses more on recommending content that is similar to the query item. 

For evaluation, we use metric that we call HIT@3.
The HIT@3 metric evaluates the performance of a ranking model by assessing whether the top three recommended items in a group of items recommended at the same time received corresponding user action, e.g., HIT@3 of Save evaluates on average how many top three recommended items by the model received Save action. A higher HIT@3 score indicates better model performance for positive actions, whereas a lower score is preferable for negative actions like Hide, as described in \cite{xia2023transact}. In all the results, we report relative lift in HIT@3 despite the action being positive or negative.

\textbf{Input sequence construction during fine-tuning:} There are different ways to construct the input sequence being passed into the \pinfm transformer and to extract output embeddings, which are then used in feature crossing. Specifically, we compare the following variations with the HF ranking model without \pinfm:

\begin{itemize}
\item \pinfm-base: take the transformer output for the candidate item and the pretrained embedding for the candidate item, use these two embeddings for feature crossing. This is the exact version described in Figure \ref{fig:frm_diagram}

\item \pinfm-GraphSAGE: on top of the \pinfm-base, sum candidate GraphSAGE \cite{Pinsage18} embedding with the pretrained id embedding as the representation of the candidate item in transformer input. 

\item \pinfm-GraphSAGE-LT: on top of \pinfm-GraphSAGE,  add a learnable token to the sequence before candidate embedding. Use the transformer output of the candidate item, the learnable token, and the pretrained embedding for the candidate item for feature crossing.

\item \pinfm-lite-mean: without inserting the candidate item into sequence, simply mean pool the user sequence output and add this embedding along with the candidate item embedding as input and cross with other input features later in the model.

\item \pinfm-lite-last: in \pinfm-lite-mean, replace mean pooling with selecting the last token.

\end{itemize}

The comparison of aforementioned variants on HF and I2I ranking models is shown in Table \ref{tab:input_seq}. We can see a clear trade-off between metric improvements and the complexity of the model. \pinfm-GraphSAGE-LT performs the best and it is also the most complex variant. On the contrary, \pinfm-lite groups have the least metric gains but they require no inclusion of the candidate item in transformer and thus benefit from vastly improved serving efficiency.

\begin{table}[h]
\centering
\captionsetup{belowskip=-5pt}
\captionsetup{aboveskip=10pt}
\caption{Percentage change in fine-tuning ranking model Save HIT@3 with different input sequence.}
\label{tab:input_seq}
\begin{tabular}{lccc}
\toprule
Input seq & HF & I2I \\
\midrule
w/o \pinfm & 0.0\% & 0.0\% \\
\pinfm-base & 2.91\% & 1.76\% \\
\pinfm-GraphSAGE & 3.08\% & 1.92\% \\
\pinfm-GraphSAGE-LT & 3.76\% & 1.92\% \\
\pinfm-lite-mean & 1.87\% & 1.53\% \\
\pinfm-lite-last & 1.93\% & 1.49\% \\
\bottomrule
\end{tabular}
\end{table}

\textbf{Candidate item representation in fine-tuning for cold start:} To improve cold start performance of candidate items with few engagements, we find a variety of regularization and special representation techniques to be important. We compare the effectiveness of these techniques on HF ranking model in Table \ref{tab:coldstart} and different variants are detailed below:

\begin{itemize}
\item \pinfm-cs-none: the basic version of \pinfm with no special handling of item freshness in \pinfm.

\item \pinfm-cs-CIR: \pinfm-cs-none with Candidate Item Randomization (CIR), meaning 10\% of candidate item ids are randomized during training before entering embedding lookup.

\item \pinfm-cs-CIR-IDD: \pinfm-cs-CPR with Item-age Dependent Dropout (IDD), where we add item age dependent dropout. More specifically, a dropout layer with $p=0.7$ is added to \pinfm produced embeddings when candidate item age is less than 7 days, and a dropout layer with $p=0.5$ is applied to \pinfm produced embeddings when item age is between 28 days and 7 days.

\item \pinfm-cs-CIR-IDD-GSLT: \pinfm-cs-CIR-IDD with summing candidate GraphSAGE and adding a learnable token (GSLT) in sequence, this becomes the same as \pinfm-GraphSAGE-LT in Table \ref{tab:input_seq}.
\end{itemize}

To measure the cold start performance, we measure the HIT@3 for Saves in HomeFeed toward the items that are fresher than 28 days old (HF 28d) and than 7 days old (HF 7d). We can see that the techniques help significantly, turning HF 28d Save metric from around -4\% to 17\%, achieving more than 20\% boost.

\begin{table}[h]
\centering
\captionsetup{belowskip=-5pt}
\captionsetup{aboveskip=10pt}
\caption{Percentage change in HF ranking model Save HIT@3 with different cold start techniques.}
\begin{tabular}{lccc}
\toprule
Candidate Item Repr. & HF Overall & HF 28d & HF 7d \\
\midrule
w/o \pinfm & 0.0\% & 0.0\% & 0.0\% \\
\pinfm-cs-none & 3.36\% & -4.40\% & -23.98\% \\
\pinfm-cs-CIR & 3.43\% & 1.25\% & -4.38\% \\
\pinfm-cs-CIR-IDD & 3.49\% & 10.71\% & 8.16\% \\
\pinfm-cs-CIR-IDD-GSLT & 3.76\% & 17.72\% & 12.01\% \\
\bottomrule
\end{tabular}
\label{tab:coldstart}
\end{table}

\textbf{Pretraining and fine-tuning losses:}
Model performance of different settings of pretraining and fine-tuning losses evaluated on HF ranking model is shown in Table \ref{tab:loss}. 
It can be seen that Save metric keep increasing as we add more losses, Hide metric improves when $\mathcal{L}_{mtl}$ is added but deteriorates when $\mathcal{L}_{ftl}$ is added. 

The effects of the sequence losses in fine-tuning are also shown in Table 
\ref{tab:loss}. We can see that removing $\mathcal{L}_{ntl}$ will lead to a sizable drop in Save with the benefit of decreasing Hide, while further adding $\mathcal{L}_{mtl}$ can provide tiny increase in Save and recover Hide. However, adding $\mathcal{L}_{mtl}$ will further increase GPU memory usage and slow down training, and thus, we do not include $\mathcal{L}_{mtl}$ by default in fine-tuning.

\begin{table}[h]
\centering
\captionsetup{belowskip=-5pt}
\captionsetup{aboveskip=10pt}
\caption{Percentage change in HIT@3 when changing losses, baseline is using $\mathcal{L}_{ntl}$ for both pretraining and fine-tuning.}
\label{tab:loss}
\begin{tabular}{lccc}
\toprule
pretraining Loss & fine-tuning Loss & Save (lift) & Hide (lift) \\
\midrule
$\mathcal{L}_{ntl}$ & $\mathcal{L}_{ntl}$ & 0.0\% & 0.0\% \\
$\mathcal{L}_{ntl} + \mathcal{L}_{mtl}$ & $\mathcal{L}_{ntl}$ & 0.42\% & -1.27\% \\
$\mathcal{L}_{ntl} + \mathcal{L}_{mtl} + \mathcal{L}_{ftl}$ &$\mathcal{L}_{ntl}$   & 0.95\% & 2.43\% \\
$\mathcal{L}_{ntl} + \mathcal{L}_{mtl} + \mathcal{L}_{ftl}$ &none  & 0.41\% & 1.06\% \\
$\mathcal{L}_{ntl} + \mathcal{L}_{mtl} + \mathcal{L}_{ftl}$ & $\mathcal{L}_{ntl} + \mathcal{L}_{mtl} $ & 1.01\% & 1.03\% \\
\bottomrule
\end{tabular}
\end{table}



\textbf{Selection of positive actions:} As introduced in Section \ref{sec:method}, the objective of our training losses is to predict positively engaged items in the future. In such a framework, the definition of positive action could affect the quality of the pretrained model. In Table \ref{tab:pre_action}, we ablate positive action types in pretraining, evaluated by metrics on the downstream HF ranking model. As can be seen in the table, adding Download or Clickthrough action as positive actions can both individually provide positive impact, and adding all the actions but Hide and Clickthrough can lead to significant gain on Hide metric. However, adding all actions but Hide will hurt Save metric. The results reflect that selecting positive actions optimally is a non-trivial problem.

\textbf{Iterations in pretraining:} We ablate how the number of iterations during pretraining impacts the performance of HF ranking and summarize the results in Figure \ref{fig:pretrain_steps}. We test from 0 iteration to 640k iterations, with 0 iteration (no pretraining) as the baseline. We observe that as we pretrain longer, the Save and Hide both get better, though not strictly monotonically. Note that the epoch boundaries are marked by yellow vertical lines. We chose to pretrain around 3 epochs to balance pretraining cost and performance. Another interesting observation is the pretraining does not suffer from one epoch overfitting issue, which is often observed in training id-based ranking models.
{\captionsetup{belowskip=-10pt}
\begin{figure}
    \centering
    \includegraphics[width=\linewidth]{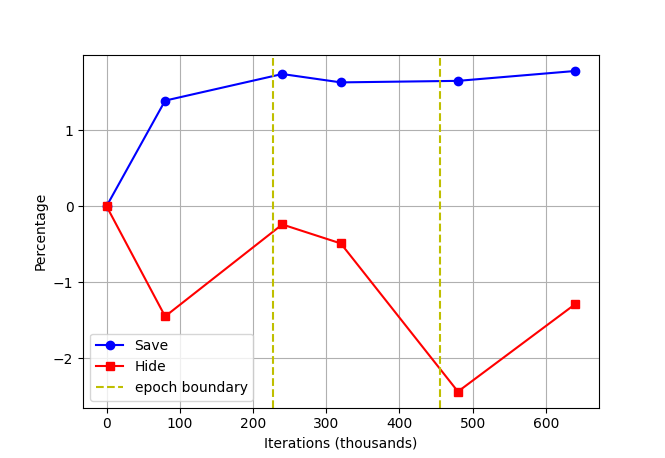}
    \caption{Save and Hide vs. number of pretraining iterations}
    \label{fig:pretrain_steps}
\end{figure}
}

\begin{table}[h]
\centering
\captionsetup{belowskip=-2pt}
\captionsetup{aboveskip=10pt}
\caption{Percentage change in HIT@3 when changing pretraining actions, with using Save as baseline.}
\label{tab:pre_action}
\begin{tabular}{lccc}
\toprule
Action types & Save (lift) & Hide (lift) \\
\midrule
Save& 0.0\% & 0.0\% \\
Save + Download & 0.21\% & -1.63\% \\
Save + Clickthrough & 0.02\% & -2.67\% \\
All - Hide & -0.2\% & -1.76\% \\
All - Hide - Clickthrough & 0.19\% & -4.1\% \\
\bottomrule
\end{tabular}
\end{table}

\textbf{Importance of fine-tuning:} In Table \ref{tab:has_finetuning}, we ablate the importance of \pinfm fine-tuning in HF ranking. In LLMs, next token prediction-based pretraining helps the model memorize world knowledge, but doesn't work out of the box for downstream applications. Fine-tuning is needed for capabilities like instruction following. Similarly, we want to understand if fine-tuning is as important in our foundation model. Experiment results show that without fine-tuning, the lift on Save diminishes to almost none, and Hide is even negatively impacted. This demonstrates that fine-tuning plays a critical role. And this shows the necessity of applying fine-tuning in both HF and I2I surfaces.

\begin{table}[h]
\centering
\captionsetup{belowskip=-2pt}
\captionsetup{aboveskip=10pt}
\caption{Percentage change in HIT@3 on HF when adding the \pinfm w/ and w/o fine-tuning, with no \pinfm as baseline}
\label{tab:has_finetuning}
\begin{tabular}{lccc}
\toprule
fine-tuning setting & Save (lift) & Hide (lift) \\
\midrule
w/o \pinfm & 0.0\% & 0.0\% \\
w/o fine-tuning (freeze \pinfm) & 0.10\% & 2.56\% \\
w/ fine-tuning & 3.76\% & -2.77\% \\
\bottomrule
\end{tabular}
\end{table}


\textbf{Choice of vocabulary size:} In Table \ref{tab:vocab_size}, we ablate the effect of scaling the vocabulary size. We observed a steady increase in model performance as we scaled the embedding vocabulary size from 20 million rows to 160 million rows. 
\begin{table}[h]
\centering
\captionsetup{belowskip=-2pt}
\captionsetup{aboveskip=10pt}
\caption{Percentage change in Save HIT@3 on HF ranking model at different vocabulary sizes, with 20M as baseline.}
\label{tab:vocab_size}
\begin{tabular}{lccc}
\toprule
Embedding vocabulary size & Save (lift) & Hide (lift) \\
\midrule
20M & 0.0\% & 0.0 \% \\
40M & 0.81\% & -0.34\% \\
80M & 0.91\% & 0.31\% \\
160M & 1.98\% & 0.92\% \\

\bottomrule
\end{tabular}
\end{table}


\subsection{Online results}

\begin{table}[H]
    \centering
    \captionsetup{belowskip=-5pt}
    \caption{Online improvements from A/B test for \pinfm. All results are statistically significant with a 95\% confidence level.}
    \label{tab:ab_results1}
    \begin{tabular}{lccc}
        \toprule
        Metric & HF Ranking &  I2I Ranking\\
        \midrule
        Sitewide Saves & \textbf{+1.20\%}  & \textbf{+0.72\%} \\
        Surface Saves & \textbf{+2.60\%}  & \textbf{+2.09\%} \\
        Fresh Saves & \textbf{+5.70\%}  & \textbf{-0.82\%} \\
        \bottomrule
    \end{tabular}
    \label{tab:online}
\end{table}

We added \pinfm to both Home Feed ranking and I2I ranking models and conducted A/B tests against the existing ranking models (Table~\ref{tab:online}). Although the baselines utilize transformers to encode user sequences~\cite{Transact23}, the number of parameters in the transformer --- combining both the neural network layers and the embedding table --- is significantly smaller, comprising less than 0.2\% parameters compared to \pinfm.


Note that \pinfm in I2I was without the cold start remediation techniques in Table~\ref{tab:coldstart}, because I2I's fresh Save metric drop was within an acceptable range. On the other hand, \pinfm in HF showed about -5\% Fresh Saves and thus we added the techniques in Table~\ref{tab:coldstart}, leading to about +10\% Fresh Saves lift.

In addition, we observed increase an in feed diversity for both HF and I2I, which is measured by the number of engaged items from topics that a user have not recently impressed or engaged with. This demonstrated the effectiveness of \pinfm to boost both engagement and feed diversity.